\documentclass[runningheads]{llncs}

\usepackage{dsbda}

\usepackage{amsmath, amsfonts}
\usepackage[english]{babel}
\usepackage{booktabs}
\usepackage{breqn}
\usepackage{caption}
\usepackage{enumitem}
\usepackage{graphicx}
\usepackage{hyperref}

\usepackage{multirow}
\usepackage{pdfpages}
\usepackage{tabularx}
\newcolumntype{R}{>{\raggedleft\arraybackslash}X}
\usepackage{xcolor}
\usepackage{longtable}

\usepackage{orcidlink}

\begin{document}

\title{Fine-Tuning Language Models for \\ Scientific Writing Support}

\author{Justin Mücke \and 
Daria Waldow \and 
Luise Metzger \orcidlink{0000-0002-5202-507X} \and 
Philipp Schauz \and
Marcel Hoffman \orcidlink{0000-0001-8061-9396} \and 
Nicolas Lell \orcidlink{0000-0002-6079-6480} \and
Ansgar Scherp \orcidlink{0000-0002-2653-9245}}

\authorrunning{J. Mücke et al.}

\institute{Universität Ulm, Germany
\email{firstname.lastname@uni-ulm.de}}

\maketitle              

\begin{abstract}
We support scientific writers in determining whether a written sentence is scientific, to which section it belongs, and suggest paraphrasings to improve the sentence.
Firstly, we propose a regression model trained on a corpus of scientific sentences extracted from peer-reviewed scientific papers and non-scientific text to assign a score that indicates the scientificness of a sentence. 
We investigate the effect of equations and citations on this score to test the model for potential biases.
Secondly, we create a mapping of section titles to a standard paper layout in AI and machine learning to classify a sentence to its most likely section.
We study the impact of context, \ie surrounding sentences, on the section classification performance.
Finally, we propose a paraphraser, which suggests an alternative for a given sentence that includes word substitutions, additions to the sentence, and structural changes to improve the writing style.
We train various large language models on sentences extracted from arXiv papers that were peer reviewed and published at A*, A, B, and C ranked conferences.
On the scientificness task, all models achieve an MSE smaller than $2\%$.
For the section classification, BERT outperforms WideMLP and Sci\-BERT in most cases.
We demonstrate that using context enhances the classification of a sentence, achieving up to a $90\%$ F1-score.
Although the paraphrasing models make comparatively few alterations, they produce output sentences close to the gold standard.
Large fine-tuned models such as T5 Large perform best in experiments considering various measures of difference between input sentence and gold standard.

Code is provided here: \url{https://github.com/JustinMuecke/SciSen}.

\end{abstract}

\section{Introduction}

Scientific writing is a complex task with many resources helping researchers and students write better text~\cite{WritingManual1,WritingManual2}.
A good structure and language facilitate the readers' understanding of the relevant content. 
Sentences in scientific papers can be expected to follow a certain scientific style, which is distinct from colloquial texts.
A typical structure of research papers with methodological and empirical contributions such as in AI and machine learning is the section sequence of an introduction, related work, methods, results, discussion, conclusion, and optionally an appendix~\cite{PaperStructure}.
Although these sections might vary based on writing styles and problem-specific content (\eg in machine learning literature, the methods section is often separated into method and experimental apparatus), readers expect to find certain pieces of information in certain sections.
Placing content into sections contrary to a reader's expectation makes it more difficult to find said information.
We investigate whether this structural clarity is better reflected in published papers of higher quality (\ie CORE database rankings\footnote{\url{http://portal.core.edu.au/conf-ranks/}}) compared to less prestigious publications.
Besides structural clarity, finding the best phrasing is a challenge, since a sentence with the same meaning can be phrased in many different ways.
While there are already solutions for related sub-tasks of sentence paraphrasing~\cite{unsupervised-divgan,paraphrasing-lstm-multiple,unsupervised-gpt2,unsupervised-edit-based,paraphrasing-unsupervised,paraphrasing-lstm,paraphrasing-transformer}, these are not specific to the domain of scientific papers.
Other tools like Grammarly\footnote{\url{https://app.grammarly.com/}} and ChatGPT\footnote{\url{https://openai.com/blog/chatgpt}} are limited to online use only and do not guarantee any data protection.
We propose a simple training procedure for paraphrasers to perform insertions, deletions, and modifications on the input text and apply it to state-of-the-art paraphrasers on scientific text.
In summary, our contributions are:

\begin{enumerate}[label=(\roman*)]
\item Scientificness score: We train regression models to discern non-scientific from scientific sentences by determining a sentence's scientificness.  
\item Section classification: We train multi-label classifiers to indicate to which sections a sentence belongs. 
Additionally, we investigate the impact of the context length and scientific quality (\ie CORE conference rank) of the input sentence on the classification performance. 
\item Sentence paraphrasing: We fine-tune the language models BART~\cite{BART} and T5 v1.1~\cite{t5-intro} small, base, and large.
We evaluate the sentence paraphrasing on these models as well as using Pegasus~\cite{pegasus} and GPT-2~\cite{gpt2}.

\end{enumerate}

The paper is structured as follows: 
We summarize the related work in Section~\ref{sec:relatedwork}.
The experimental apparatus is described in Section~\ref{sec:experimentalapparatus}.
The results are reported in Section~\ref{sec:results} and discussed in
Section~\ref{sec:discussion}, before we conclude.

\section{Related Work}
\label{sec:relatedwork}

We discuss the literature on language models and their capabilities on our tasks, \ie scoring, multi-label classification, and paraphrasing. 
We provide a brief overview of existing commercial tools for writing assistance to further demonstrate the relevance of this area of research.

\subsection{Pre-trained Encoder Language Models}
Encoder-only language models learn representations for each token of an input sequence.
BERT~\cite{BERT} is a encoder-only language model pre-trained using masked language modelling (MLM) and next sentence prediction (NSP).
The pre-trained model can be fine-tuned on various downstream tasks~\cite{BERT}. 
There are many variations of BERT~\cite{SciBERT,Electra,BART,RoBERTa} with Sci\-BERT~\cite{SciBERT} being the most relevant to us.
It is pre-trained on a corpus of scientific papers from bio-medicine and computer science to increase its performance in those domains~\cite{SciBERT-rerank}.
Due to the high computational cost for pre-training, ELECTRA~\cite{Electra} aims to increase pre-training efficiency. 
ELECTRA uses two neural networks, a generator which is discarded after training and a discriminator. 
The generator plausibly substitutes masked tokens from an input sentence.
The discriminator has to distinguish between tokens from the original input sequence and tokens generated by the generator. 
This way, each token of the input sequence contributes to the loss of the discriminator, instead of only the masked tokens as in BERT. 

\subsection{Pre-trained Decoder Language Models}
Decoder language models~\cite{BART,gpt2,t5-intro} are designed for text generation.
They take textual input and generate a new output sequentially token by token. 
Auto-regressive decoders use already generated tokens to generate the following tokens~\cite{BART,gpt2,t5-intro}.

We use four language models for paraphrasing, namely BART~\cite{BART}, 
Pegasus~\cite{pegasus}, 
T5 v1.1~\cite{t5-intro}, 
and GPT-2~\cite{gpt2}.
BART~\cite{BART} is a general-purpose sequence-to-sequence model that adds a left-to-right auto-regressive decoder to BERT.
Pegasus~\cite{pegasus} is a sequence-to-sequence language model trained by gap-sentence generation, which is comparable to MLM, but masks whole sentences instead of words.
We use a variant of Pegasus fine-tuned on paraphrasing~\cite{PegaPara}.
The encoder-decoder model T5~\cite{t5-intro} introduces the concept of task instructions such as translation, classification, and summarization as part of the prompts.
These instructions are provided to T5 while being fine-tuned on multiple tasks at the same time.
We use a version of T5 that is yet not fine-tuned on multiple tasks, \ie does not provide a token-based task execution.
Instead, we fine-tune our T5 (in version 1.1) on the task of paraphrasing as this is the only task we want to perform.
Thus, we omit using task-specific prefixes during fine-tuning.
GPT-2~\cite{gpt2} is a decoder-only model and is trained on the next token prediction objective. 
It generates text in an auto-regressive manner to continue the prompt.

\subsection{Text Classification}
A classification task can be either single-label or multi-label.
In multi-label classification, a classified object can be associated with none, one, or more classes. 
BERT-based architectures achieve state-of-the-art results in many tasks, including single-label and multi-label text classification~\cite{ACL22GalkeScherp,ACL22MultiLabel}. 
For scientific texts, Sci\-BERT has been used to perform citation intent classification~\cite{CIC} and classification of paper titles and abstracts to research disciplines~\cite{TextClassML1}.
Sci\-BERT and BioBERT outperformed BERT on texts from STEM domains, but were outperformed on texts on language or history~\cite{TextClassML1}.
The performance of these models can be influenced by characteristics of the input data, \eg adding document context can improve task performance~\cite{ContextNER}.
Galke et al.~\cite{ACL22MultiLabel} showed that WideMLP~\cite{ACL22GalkeScherp}, a Multi-Layer Perceptron (MLP) model with a wide hidden layer, is a strong baseline for text classification in both the single-label and multi-label scenarios.
However, BERT achieved state-of-the-art results in text classification for various datasets.
To the best of our knowledge, there has been no attempt to classify sentences of scientific text according to their corresponding section.

\subsection{Sentence Transformation and Paraphrasing}
A common task performing sentence transformations is Neural Machine Translation (NMT)~\cite{DBLP:journals/csur/DabreCK20}.
Many approaches for machine translation require large amounts of training data~\cite{paraphrasing-lstm-multiple,paraphrasing-lstm,paraphrasing-transformer}, with transformers achieving state-of-the-art performance~\cite{paraphrasing-transformer}.
Besides translation, a text can be transformed by paraphrasing, which changes an existing sentence while preserving its meaning~\cite{paraphrasing_definition_from_upsa2,paraphrasing_definition_from_upsa1,paraphrasing_definition_from_upsa3}. 
Many paraphrasing approaches are limited to word-level changes~\cite{unsupervised-edit-based,paraphrasing-unsupervised}. 
Rudnichenko et al.~\cite{text_data_rephrasing} propose a system that paraphrases individual sentences, including changes to the word order.  
These sentence transformation methods are all supervised, \ie the training datasets have a parallel corpus containing two versions of each sentence. 
Such datasets are expensive to create.
To tackle this challenge, unsupervised paraphrasing approaches create training data by inserting, replacing, or deleting words from a sentence~\cite{unsupervised-divgan,unsupervised-gpt2,unsupervised-edit-based,paraphrasing-unsupervised}.
Other approaches create multiple alternative sentences~\cite{paraphrasing-lstm-multiple,unsupervised-edit-based} and apply evaluation methods on each suggestion.
A special case of paraphrasing is text style transfer (TST) which aims to change the style of a text to imitate a specific writing style~\cite{tst_evaluation,tst_survey}.

\subsection{Tools to Improve Writing Quality}
There are various tools to assess writing quality, which target spelling mistakes, grammar errors, long sentences, and suggest paraphrases. 
Specifically, Writefull\footnote{\url{https://www.writefull.com/}} is a tool for scientific writing.
It allows sentence paraphrasing and is trained on scientific text, which sets it apart from tools for general English language. 
QuillBot\footnote{\url{https://quillbot.com/}} provides paraphrasing for general writing. 
LanguageTool\footnote{\url{https://languagetool.org/}} and Grammarly\footnote{\url{https://www.grammarly.com/}} provide general spelling and grammar improvements. 
However, even if a sentence is grammatically, orthographically, and semantically correct, it could still be non-scientific in style.  
Recent developments suggest that these tasks can be tackled by tools like ChatGPT (based on InstructGPT~\cite{InstructGPT}).
However, it cannot be used offline, is expensive to run, and does not guarantee data protection.
Thus, we train large language models ourselves to perform paraphrasing tasks on a local infrastructure.

\section{Experimental Apparatus}
\label{sec:experimentalapparatus}
In this section, we describe the datasets, the preprocessing, the procedure for each task, the hyperparameter search, and the evaluation measures.

\subsection{Datasets}
\label{sec:datasets}
We use papers published on arXiv until May $2022$ with LaTeX available that were accepted at A*, A, B, and C ranked conferences of the Australian CORE2021 database.
To map papers with their respective conferences, we use the Papers With Code database\footnote{\url{https://production-media.paperswithcode.com/about/papers-with-abstracts.json.gz}}.
Since we extract the structure of the papers, we drop all papers that are not using any \texttt{\textbackslash{}section\{...\}} command in \LaTeX.
Overall, we have a total of $26,201$ papers, of which $21,774$ are from A*, $3,665$ from A, $530$ from B, and $232$ from C-ranked conferences.

For the scientificness score task, we complement our arXiv text with non-scientific sentences from Reddit comments\footnote{\url{https://files.pushshift.io/reddit/comments/}}, sci-fi stories\footnote{\url{https://www.kaggle.com/datasets/jannesklaas/scifi-stories-text-corpus}}, and subsets of different Twitter datasets~\cite{UkraineTweets,CovidTweets}.
For the section task, we can use our arXiv dataset as-is.
Finally, for paraphrasing, we create two parallel datasets by reducing the quality of the sentences, \eg replacing words with colloquial synonyms. 
The first dataset is Pegasus-DS, which is created by changing sentences using Pegasus fine-tuned for paraphrasing~\cite{pegasus}. 
The second dataset IDM-DS is created by randomly inserting, deleting, and modifying up to half of the tokens of each sentence based on MLM using BERT~\cite{BERT}.
To evaluate the paraphraser, we additionally use Grammarly’s Yahoo Answers Formality Corpus (GYAFC)~\cite{yahoo-dataset} for testing. 
GYAFC contains informal and formal sentences with four human-written paraphrases. 
We use $1,332$ sentences from the category family and relationships, as the dataset provides output sentences from other models in this category. 
The statistics of the datasets are summarized in Table~\ref{tab:senStats}.

\begin{table}[th]
\centering        
\caption{The number of sentences in the datasets and the sentences removed by applying filters.
The filters remove sentences with non-ASCII characters, minimum length threshold, maximum length threshold, and if they contained a non-capitalized first character or did not end with a punctuation.}
\footnotesize
\begin{tabular}{l|r|rrrrr|r}
\toprule
Dataset name                          & Number     & \multicolumn{5}{c|}{Filter}    & Remaining     \\
                                      &       & ASCII & Short     & Long    & First     & Last     &      \\
\midrule
arXiv                      &$ 5,283,451 $&$ 51,201    $&$ 61,705    $&$ 1,905   $&$ 197,081   $&$ 12,696  $&$ 4,958,863  $\\
w. section ID &$ 2,864,755 $&$ 27,357    $&$ 32,110    $&$ 790     $&$ 110,467   $&$ 6,667   $&$ 2,687,364   $\\
\midrule
Books                                 & $1,763,465 $&$  0        $&$ 149,215   $&$ 1,006   $&$ 10,673    $&$ 0       $&$ 1,613,244  $\\
Reddit                                &$ 279,288   $&$  11,774   $&$ 51,582    $&$ 340     $&$ 5,638     $&$ 0       $&$ 217,225     $\\
Twitter                               &$ 268,419   $&$  233,272  $&$ 241       $&$ 9       $&$ 0         $&$ 0       $&$ 35,108  $\\     \bottomrule
\end{tabular}

\label{tab:senStats}
\end{table}

\subsection{Preprocessing}
\label{sec:preprocessing}
Citations and references were replaced by a \texttt{<reference>}-token.
In the case of \texttt{\textbackslash{}citeauthor}, \texttt{\textbackslash{}citet}, etc., which produces author names in the \LaTeX{} output, we insert a random name\footnote{\url{https://www.kaggle.com/datasets/jojo1000/facebook-last-names-with-count}} to preserve the structure of the sentence. 
Math syntax was replaced by \texttt{<equation>}-tokens.
For the section classifier, we remove all sections with titles that cannot be mapped to one of our predefined categories, \ie the classes our models are trained on.
These classes are ``introduction'', ``related work'', ``method'', ``experiment'', ``result'', ``discussion'', and ``conclusion''.
Section titles extracted from the papers that fall into more than one category are mapped to all of the categories they consist of.
For example, the paper section entitled Introduction and Background is mapped internally to the two classes ``introduction'' and ``related work''. 
Our corpus includes machine learning papers containing \texttt{[MASK]} as a word.
Since the insert, delete, and modify (IDM) process recognizes \texttt{[MASK]} as a special input token, we removed the brackets in the IDM-DS dataset.

We split the input at end-of-sentence punctuation symbols \texttt{.}, \texttt{?}, and \texttt{!} to obtain sentences.
As documented in Table~\ref{tab:senStats}, we drop sentences containing non-ASCII characters to ensure that classification tasks are not trivial due to emojis or similar characters. 
We limit the length of extracted sentences to be at least $4$ and at most $100$ words.
The upper limit was set as five times the average sentence-length in non-fiction writing as well as five times the highest recommended sentence length in English writing~\cite{rudnicka2018variation}.
We filter sentences that do not follow basic orthography, \ie that do not start with a capital letter or end with end-of-sentence punctuation. 

\subsection{Procedure}
\label{sec:procedure}

\subsubsection{Scientificness Score}
\label{sec:procedure_Score}

We then fine-tune BERT base~\cite{BERT} and Sci\-BERT~\cite{SciBERT} and train a Bag-of-Words WideMLP~\cite{ACL22MultiLabel} with one hidden layer from scratch to predict the scientificness score.
This is interpreted as a regression score, where we assign a score of $0.9$ to scientific sentences and $0.1$ to non-scientific sentences during training.
We evaluate whether the conference rank of the paper affects the models' scores.

Furthermore, we investigate the effect of using the \texttt{<equation>} and \texttt{<re\-fer\-ence>} tokens by separately evaluating scientific sentences with and without such tokens.
We also add these tokens to $100,000$ randomly sampled sentences from the Books dataset and compare the scores of the modified (\ie with tokens ingested) and original sentences.
            
\subsubsection{Section classification}
\label{sec:procedure_MLSC}
We use BERT base~\cite{BERT}, Sci\-BERT~\cite{SciBERT}, and a Bag-of-Words WideMLP~\cite{ACL22MultiLabel,ACL22GalkeScherp} with one hidden layer. 
Since a sentence might have multiple section labels, we train the models as multi-label classifiers.
We examine the influence of the amount of context provided as input to the model by varying the context length in training and testing.
The input contexts provided to the models are a single-sentence, two sentences, and three sentences (up to BERT's maximum input length of $512$ tokens). 
Two-sentence input contains the sentence of interest plus its predecessor, and three-sentence input contains the sentence of interest plus its predecessor and successor. 
Additionally, we examine the influence of the conference rank on classification performance, \ie we separately evaluate sentences from conferences ranked as A*, A, and B and C combined.
Thus, papers from B and C are treated as one bucket, since the number of C papers ($232$ publications) is small.

\subsubsection{Sentence paraphrasing}
\label{sec:procedure_Rephrasing}
The training of the paraphrasing models is based only on text from A* and A conference papers to ensure high-quality training data. 
The models are fine-tuned on Pegasus-DS and IDM-DS to reconstruct the original scientific sentence from the corrupted version.
We fine-tune models based on T5 v1.1 in the variants small, base, and large, and BART base.
We use GPT-2 with the prompt prefix ``In scientific language, '' and include identity as a baseline.
For all models, we apply beam search with a width of $5$ to generate paraphrases and select the one with the highest probability.

The metrics are computed on the test split of each dataset.
The test splits are divided into buckets which reflect the amount of changes made compared to the gold standard relative to the sentence length.
The changes range from $0\%$ to $50\%$ in $10\%$ steps resulting in six buckets, where, for example, $40\%$ means that $6$ words are changed in a $15$ words long sentence.
For IDM-DS, the number of changes is known from creating the dataset.
Thus, we control the amount of changes in the sentences, but it may happen in an unlikely case that a sequence of operations could undo an earlier change on a sentence, \eg an insert followed by a delete later on.
For Pegasus-DS, we use the word error rate~\cite{WER} (WER) between the original and corrupted sentence to measure the amount of changes.
WER is a word-level version of the edit distance representing the number of substitutions, deletions, and insertions divided by the original sequence length.

Additionally, we evaluate the performance of our models on the GYAFC dataset to assess their capabilities to transform text from informal to formal writing.
We compare our models to the results of the GYAFC paper~\cite{yahoo-dataset}, which includes a non-scientific paraphraser, a rule-based approach, and a NMT-based model combined with rules (denoted as NMT combined).
We also compare to the results of two text-style transfer models, DualRL and DAST-C~\cite{tst_evaluation}.

\subsection{Hyperparameter Optimization}
\label{apdx:hyperparameter-tuning}

For all tasks and datasets, we use random $70$:$20$:$10$ train, validate, and test split.
We tune hyperparameters on a $10\%$ subset of the train and validation data.

\subsubsection{Scientificness score}
For the scientificness score, we fine-tune BERT, Sci\-BERT, and WideMLP using AdamW. 
We test the learning rates $1\cdot10^{-5}$, $3\cdot10^{-5}$, $5\cdot10^{-5}$, the dropout rates $0.1$, $0.3$, and $0.5$, and the values $0.05$, $0.01$, and $0.001$ for the weight decay. 
We train BERT and Sci\-BERT for five epochs and WideMLP for ten epochs, since the loss stopped decreasing there.
We use a batch size of $8$ as this was the highest one to fit on our GPU.
Sci\-BERT performed best with a learning rate of $1\cdot 10^{-5}$, $0.3$ dropout rate, and $0.1$ weight decay. 
BERT performed best using a learning rate of $1\cdot 10^{-5}$, $0.1$ dropout rate, and $0.5$ weight decay. 
WideMLP performed best with a learning rate of $0.05$, $0.3$ dropout rate, and $0.05$ weight decay.

\subsubsection{Section classification}
We use Adam for fine-tuning BERT and Sci\-BERT for multi-label classification.
We set a maximum of $15$ epochs with early stopping if the validation loss did not decrease for two epochs. 
Since the models stopped improving after $1$ to $3$ epochs, we did not tune the number of epochs further. 
We train with a batch size of $32$, which was the maximum that reliably fit on our GPU. 
We experimented with learning rates of $1\cdot10^{-5},3\cdot10^{-5},5\cdot10^{-5}$ and with $\lambda$ thresholds of $0.5$, $0.3$, $0.2$, and $0.1$, \ie the threshold above which a label is assigned in multi-label classification. 
The best-performing parameters were found to be a learning rate of $1\cdot10^{-5}$ and $\lambda = 0.2$ for all transformer models.
For WideMLP, we use AdamW and train for $100$ epochs with a learning rate of $10^{-1}$ for all datasets, following Galke et al.~\cite{ACL22MultiLabel}.
After testing the $\lambda$ thresholds, we achieved the best results with $\lambda = 0.2$.

\subsubsection{Sentence paraphrasing}
We use AdamW to fine-tune T5 and BART.
Performance stopped improving after three to five epochs, so we set the number of epochs to $5$.
As we did not observe a performance impact of changing the batch size, we use the highest batch size that fits on our GPU, which was between $20$ to $200$ depending on the model. 
We use a learning rate of $2\cdot10^{-5}$ for the experiments.
We test the values of $0.05$, $0.01$, and $0.001$ for weight decay.
Different metrics favored different values, so we use $0.001$ as  weight decay because the models perform consistently well for all metrics using this value.

\subsection{Measures}
\label{sec:measures}

Since the scientificness score is a regression task, we evaluate it using mean squared error (MSE).
For the section classification, we use sample-based F1, following Galke et al.~\cite{ACL22MultiLabel}. 
For the sentence paraphrasing, BLEU, METEOR, and BERTScore measure the difference to the gold standard and self-BLEU measures the difference to the input. 
BLEU calculates $n$-gram similarity with $n=4$ and is the standard metric for paraphrasing~\cite{unsupervised-gpt2,tst_evaluation,unsupervised-edit-based,paraphrasing-unsupervised,yahoo-dataset}.
METEOR is similar to BLEU but includes synonym matching to better match human judgements~\cite{unsupervised-gpt2}.
BERTScore~\cite{BERTscore} measures semantic changes by calculating the cosine similarity of sentence embeddings of two sentences~\cite{unsupervised-divgan,tst_evaluation,unsupervised-edit-based}.
We use Sci\-BERT~\cite{SciBERT} to generate these embeddings, since we apply the score on scientific text.
The self-BLEU calculates a BLEU score between the input sentence and output sentence and is a measure of the amount of changes done by each model.

\section{Results}
\label{sec:results}

For the scientificness score task, we achieved an MSE of $0.181\%$ for the fine-tuned BERT,  $0.213\%$ for fine-tuned Sci\-BERT, and $0.049\%$ for the best performing WideMLP. 
The results of our study of the effect of \texttt{<equation>} or \texttt{<reference>} tokens on the models are presented in Table~\ref{tab:sciscore_token}.
For scientific text, the score is roughly the same for sentences with and without such tokens.
However, the standard deviation with tokens is three orders of magnitude lower for sentences with the tokens.
Adding such tokens to non-scientific text pushes the score towards more scientificness and also increases the standard deviation.

\begin{table}
\centering
\caption{Scientificness score of sentences grouped by conference rank. 
Left: Only sentences without \texttt{<equation>} or \texttt{<reference>} tokens. 
Right: Only sentences with such tokens are evaluated. We also report scores for non-scientific sentences (NSC) and modified-NSC (m-NSC), where the equation and reference tokens were artificially inserted at random.}
\label{tab:sciscore_token}
\footnotesize
\resizebox{.99\textwidth}{!}{
\begin{tabular}{lc|rrrrr|rrrrr}\toprule
&& \multicolumn{5}{c|}{Without equation and reference tokens} & \multicolumn{5}{c}{With equation and reference tokens} \\
     Model & MSE & A* & A & B & C & NSC & A* & A & B & C & m-NSC    \\ \toprule
    \multirow{2}{*}{BERT}    & Avg.  &$ .8993 $&$ .8985 $&$ .8984 $&$ .8918 $&$.1054$&$ .9016 $&$ .9016 $&$ .9016 $&$ .9016 $&  $.8142$\\
                             & SD    &$ .0392 $&$ .0449 $&$ .0450 $&$ .0804 $&$.0786$&$ .0001 $&$ .0001 $&$.0001 $&$.0001 $&  $.2334$\\\midrule
    \multirow{2}{*}{Sci\-BERT} & Avg.&$ .9004 $&$ .8988 $&$ .8992 $&$ .8892 $&$.1034$&$ .9032 $&$ .9032 $&$ .9032 $&$ .9031 $&$ .8819 $ \\
                             & SD    &$ .0438 $&$ .0548 $&$ .0514 $&$ .0977 $&$.0778$&$ .0000 $&$ .0000 $&$ .0000 $&$ .0000 $&  $.1096$\\\midrule
    \multirow{2}{*}{WideMLP} & Avg. &$ .8914 $&$ .8880 $&$ .8889 $&$ .8645 $&$.1388$&$ .8913 $&$ .8878 $&$ .8885 $&$ .8648 $&  $.5168$\\
                             & SD    &$ .0522 $&$ .0617 $&$ .0611 $&$ .0997 $&$.1197$&$ .0523 $&$ .0615 $&$ .0606 $&$ .0988 $&  $.1387$\\\bottomrule
\end{tabular}
}
\end{table}

For the section classification, the best sample-based F1-score was achieved by BERT trained on a three-sentence input taken from conferences ranked A*. 
See Table \ref{tab:section-results} for detailed results.
Table \ref{tab:section-results-context} shows the results for the context length experiment.
In this setting, the BERT model trained on two and evaluated on three sentences achieved the best performance.

\begin{table}
\centering
\caption{Sample-based F1-score (in \%) on section classification. 
Model trained on all data with different context sizes and evaluated per conference level.  
1-sentence input uses the current sentence only, 2-sentence additionally considers the previous, and 3-sentence additionally the previous and next sentence.
}
\label{tab:section-results}
\footnotesize
\begin{tabular}{ll|>{$}r<{$}>{$}r<{$}>{$}r<{$}>{$}r<{$}}
\toprule
    Input & Model  & $all$ & $A*$ & $A$ & $B/C$ \\
    \midrule
    \multirow{3}{*}{1-sentence} & BERT & 68.37 & 68.97 & 64.69 & 64.60 \\
    & Sci\-BERT & 68.68 & 69.30 & 64.96 & 64.42 \\ 
    & WideMLP & 40.97 & 41.42 & 38.20 & 38.61 \\
    \midrule      
    \multirow{3}{*}{2-sentences} & BERT & 79.40 & 79.77 & 77.05 & 77.04 \\
    & Sci\-BERT & 79.16 & 79.58 & 76.62 & 76.20 \\
    & WideMLP & 60.36 & 61.26 & 54.73 & 54.58 \\     
    \midrule
    \multirow{3}{*}{3-sentences} & BERT & \textbf{90.10}  & \textbf{90.26} & \textbf{89.13} & \textbf{88.86} \\
    & Sci\-BERT & 88.87 & 89.05 & 87.67 & 87.54 \\
    & WideMLP & 67.43 & 68.37 & 61.30 & 62.05 \\
    \bottomrule
\end{tabular}
\end{table}

\begin{table}
\centering
\caption{Sample-based F1-score (in \%) of the section classification task from papers of all ranks. Context (Train) indicates the context during training, while Context (Eval) refers to the context for evaluation.
1-sentence input uses the current sentence only, 2-sentence additionally considers the previous, and 3-sentence additionally the previous and next sentence.
}
\label{tab:section-results-context}
\footnotesize
\begin{tabular}{ll|>{$}r<{$}>{$}r<{$}>{$}r<{$}}
\toprule
Context      &         & \multicolumn{3}{c}{Context (Eval)}  \\
(Train)      & Model   & $1-sentence$          & $2-sentences$      & $3-sentences$ \\ \midrule
             & BERT    & 68.37                 & 78.35              & 81.72        \\
1-sentence   & Sci\-BERT & 68.68                 & 78.81              & 81.81        \\
             & WideMLP & 40.97                 & 42.46              & 43.16        \\   \midrule
             & BERT    & 73.96                 & 79.40              & \textbf{90.30} \\
2-sentences  & Sci\-BERT & 73.05                 & 79.16              & 89.39        \\
             & WideMLP & 51.44                 & 60.36              & 64.82        \\   \midrule
             & BERT    & 72.68                 & 90.04              & 90.10        \\
3-sentences  & Sci\-BERT & 71.48                 & 88.35              & 88.87        \\
             & WideMLP & 49.51                 & 62.53              & 67.43        \\
                   \bottomrule
\end{tabular}
\end{table}

For sentence paraphrasing, the results in Table~\ref{tab:style-results-testset-bins-both} show that T5 large performed best on the fine-tuning datasets. 
On the IDM-DS, sentences are changed more (self-BLEU) than on the Pegasus-DS, and at the same time the changed sentences are closer to the gold standard (BLEU). 
On the GYAFC dataset, see results in Table~\ref{tab:style-results-gyafc}, T5 base has the highest BLEU score.
Overall, the fine-tuning on IDM-DS performed better than Pegasus-DS as the BLEU score is higher, but at the cost of a higher self-BLEU.

The BLEU and METEOR scores improve with larger model sizes, \ie the generated sentences are closer to the original sentences when larger models are used.
The BERTScore also improves for larger models, showing that the sentences' semantics is preserved better.
Increasing the model size decreases self-BLEU, \ie larger models change the input sentences to a higher degree.
On the Pegasus-DS the difference of BLEU and METEOR is quite large, while being quite small on the IDM-DS.
This means that on IDM-DS, the models have a higher chance of replacing words with the correct synonyms, while the paraphrasing output on the Pegasus-DS remains to have larger differences to the gold standard.
The models fine-tuned on Pegasus-DS have a higher self-BLEU than the IDM-DS models.
Therefore, the models' inputs are closer to their outputs, \ie these models make fewer changes on average.
Table~\ref{tab:style-results-gyafc} shows that our models have higher BLEU and self-BLEU scores on the GYAFC dataset, \ie our models make fewer changes to the input sentences and still produce outputs close to the gold standard.

\begin{table}
\centering
\caption{Results (in \%) for Pegasus-DS (left) / IDM-DS (right) divided into buckets based on Word Error Rate (WER) and change rate (CR), respectively. All models are fine-tuned on the respective dataset. Identity returns the input.
}
\label{tab:style-results-testset-bins-both}
\footnotesize
\resizebox{.98\textwidth}{!}{
\begin{tabular}{ll|>{$}r<{$}>{$}r<{$}>{$}r<{$}>{$}r<{$}} \toprule
WER/CR & Model & $BLEU $ \uparrow  & $METEOR $ \uparrow & $BERT $ \uparrow & $sBLEU $  \\ \midrule
\multirow{5}{*}{0} & identity &  69.56/74.62 & 78.03/83.91 & \textbf{87.79}/97.41 & 100.00/100.00 \\ 
& T5 small & 69.01/84.91 & 76.91/87.66 & 86.94/90.53 & 93.50/\textbf{58.17} \\ 
& T5 base & 69.00/85.95 & 77.23/88.44 & 86.74/90.81 & 88.35/76.00 \\ 
& T5 large & \textbf{69.57}/86.85 & 78.07/89.03 & 86.86/\textbf{98.50} & \textbf{85.27}/74.93 \\ 
& BART base & 69.24/\textbf{87.71} & \textbf{78.35}/\textbf{91.23} & 87.44/91.56 & 86.11/75.44 \\
\midrule
\multirow{5}{*}{$10\%$} & identity & 48.25/67.16 & 63.05/81.53 & 82.84/96.47 & 100.00/100.00 \\
& T5 small &  48.47/80.03 & 62.56/86.17 & 81.91/89.82 & 90.15/\textbf{55.24} \\
& T5 base  &  48.83/81.64 & 63.45/87.07 & 81.54/90.24 & 80.73/71.09 \\
& T5 large &  \textbf{49.57}/83.13 & 64.78/87.78 & 81.81/\textbf{98.20} & \textbf{75.73}/69.52 \\
& BART base & 49.42/\textbf{83.72} & \textbf{65.31}/\textbf{90.09} & \textbf{82.88}/91.08 & 79.15/70.23 \\
\midrule
\multirow{5}{*}{$20\%$} & identity & 36.28/58.16 & 55.28/78.71 & 79.71/94.23 & 100.00/100.00 \\
& T5 small & 36.95/71.73 & 55.08/83.47 & 78.65/88.26 & 87.13/\textbf{49.86} \\
& T5 base  & 37.59/74.09 & 56.29/84.60 & 78.42/88.91 & 75.31/66.27 \\
& T5 large & \textbf{38.60}/76.40 & 57.60/85.58 & 78.83/\textbf{97.54} & 69.40/63.87 \\
& BART base & 38.27/\textbf{76.41} & \textbf{58.04}/\textbf{87.76} & \textbf{80.06}/89.87 & \textbf{74.22}/64.86 \\
\midrule
\multirow{5}{*}{$30\%$} & identity &  27.32/52.38 & 50.00/77.05 & 76.87/92.40 & 100.00/100.00 \\
& T5 small &  28.34/65.34 & 50.15/81.58 & 76.03/87.01 & 85.01/45.87 \\
& T5 base  &  29.50/68.41 & 51.55/82.87 & 76.15/87.87 & 71.31//63.72 \\
& T5 large &  \textbf{30.59}//\textbf{71.25} & 52.75/83.96 & 76.63/\textbf{97.04} & \textbf{64.45}/\textbf{60.80} \\
& BART base &  30.18/70.91  & \textbf{53.16}/\textbf{86.06} & \textbf{77.92}/88.87 & 69.84/62.12 \\
\midrule
\multirow{5}{*}{$40\%$} & identity & 22.02/46.58 & 47.06/75.46 & 74.86/90.43 & 100.00/100.00 \\ 
& T5 small &  23.29/59.10 & 47.88/79.65 & 74.83/85.67 & 84.62/\textbf{41.76} \\
& T5 base  &  25.16/62.61 & 49.62/81.04 & 75.28/86.71 & 67.98/60.95 \\
& T5 large &  \textbf{26.43}/\textbf{65.89} & 50.93/82.25 & 75.87/\textbf{96.47} & \textbf{60.26}/57.39 \\
& BART base & 25.82/65.18 & \textbf{51.24}/\textbf{84.30} & \textbf{76.98}/87.76 & 65.62/59.16 \\
\midrule
\multirow{5}{*}{$50\%$} & identity & \textbf{36.28}/41.28 & 55.28/73.82 & \textbf{79.71}/88.38 & 100.00/100.00 \\
& T5 small & 29.03/52.89 & 55.13/77.69 & 75.94/84.19 & 80.81/\textbf{37.63} \\
& T5 base  & 32.65/56.76 & 58.52/79.22 & 77.28/85.41 & 63.32/58.79 \\
& T5 large & 34.67/\textbf{60.40} & 60.46/80.51 & 78.19/\textbf{95.88} & \textbf{56.28}/54.73 \\
& BART base & 34.19/59.51 & \textbf{61.15}/\textbf{82.57} & 79.21/86.54 & 60.17/56.79 \\

\bottomrule
\end{tabular}
}
\end{table}

\begin{table*}[th]
\centering
\caption{Results (in \%) of our models on the GYAFC dataset. All models are evaluated with the same implementation of the metrics for either our own models (``own'', \ie we trained the models), on the models' output provided by the original papers (marked as ``output'' in the provenance column), or model weights (indicated by ``weights''). 
The best scores per metric are marked in bold.}
\label{tab:style-results-gyafc}
\footnotesize
\resizebox{.98\textwidth}{!}{

\begin{tabular}{ll|>{$}r<{$}>{$}r<{$}>{$}r<{$}>{$}r<{$}l} \toprule
    Model & Fine-tuning & $BLEU$\uparrow & $METEOR$\uparrow & $BERTScore$\uparrow  & $sBLEU$\downarrow
    & Provenance \\ 
    \midrule
    Original Informal & - & 55.01  & 20.25 & 94.00 & 100.00 
    & output~\cite{yahoo-dataset}  \\
    Rule-based    & -     & 49.49 & 17.20 & 94.39 &  57.92 
    & output~\cite{yahoo-dataset}  \\
    NMT Combined  & GYAFC & 52.50  & 17.23 & \textbf{94.93} &  47.86
    & output~\cite{yahoo-dataset}  \\
    \midrule
    DualRL & GYAFC & 39.75  & 16.93 & 92.38 & 45.99
    & output~\cite{tst_evaluation} \\
    DAST-C & GYAFC & 36.14 & 18.52 & 90.99  & 47.81  
    & output~\cite{tst_evaluation} \\
    Pegasus & - & 49.72  & 16.80 & 86.33 & 35.98  
    & weights~\cite{PegaPara} \\
    IDM   & - & 49.52  & 17.93 & 92.76 & 80.02 
    & own \\
    GPT-2 & - &  1.48  & 17.30 & 78.84 &  \textbf{1.38}  
    & own \\ 
    \midrule
    T5 small  & Pegasus-DS & 48.73  & 22.04 & 84.35 & 65.23 
    & own \\
    T5 base   & Pegasus-DS & 50.04  & \textbf{22.22} & 66.30 & 70.30 
    & own \\
    T5 large  & Pegasus-DS & 49.91 &  21.56 & 85.65 & 74.65 
    & own \\
    BART base & Pegasus-DS & 54.56 & 20.75 & 67.13 & 86.42  
    & own \\
    \midrule
    T5 small  & IDM-DS & 58.47  & 20.75 & 88.66  & 86.65
    & own \\
    T5 base   & IDM-DS & \textbf{57.21} & 20.63 & 67.62  & 90.60 
    & own \\
    T5 large  & IDM-DS & 55.23  & 20.45 & 87.89  & 91.28 
    & own \\
    BART base & IDM-DS & 54.48  & 20.42 & 67.40 & 93.24 
    & own \\
    \bottomrule
\end{tabular}
}
\end{table*}

\section{Discussion}
\label{sec:discussion}

\subsection{Key Results}
\label{sec:keyresults}

\subsubsection{Scientificness score}
All models score sentences from scientific papers at a value of around $0.9$ (see Table \ref{tab:sciscore_token}) with the highest scores provided by Sci\-BERT.
For all models, the mean output decreases, and the standard deviation increases with decreasing conference rank.
This suggests that lower-ranked conferences contain, on average, fewer scientific sentences.
Therefore, low-ranked conferences have a broader range of sentence quality and include more sentences with a lower scientificness score.

The experiment on the influence of \texttt{<equation>} and \texttt{<reference>} tokens (see Table~\ref{tab:sciscore_token}) shows that transformer models rank sentences containing such tokens higher than sentences without such tokens.
This indicates that the models have learned to connect sentences containing these tokens with higher scientificness. 
The low standard deviation indicates a more stable prediction of the scientificness score for sentences containing these tokens. 

\begin{figure}
\centering
\includegraphics[scale=0.6]{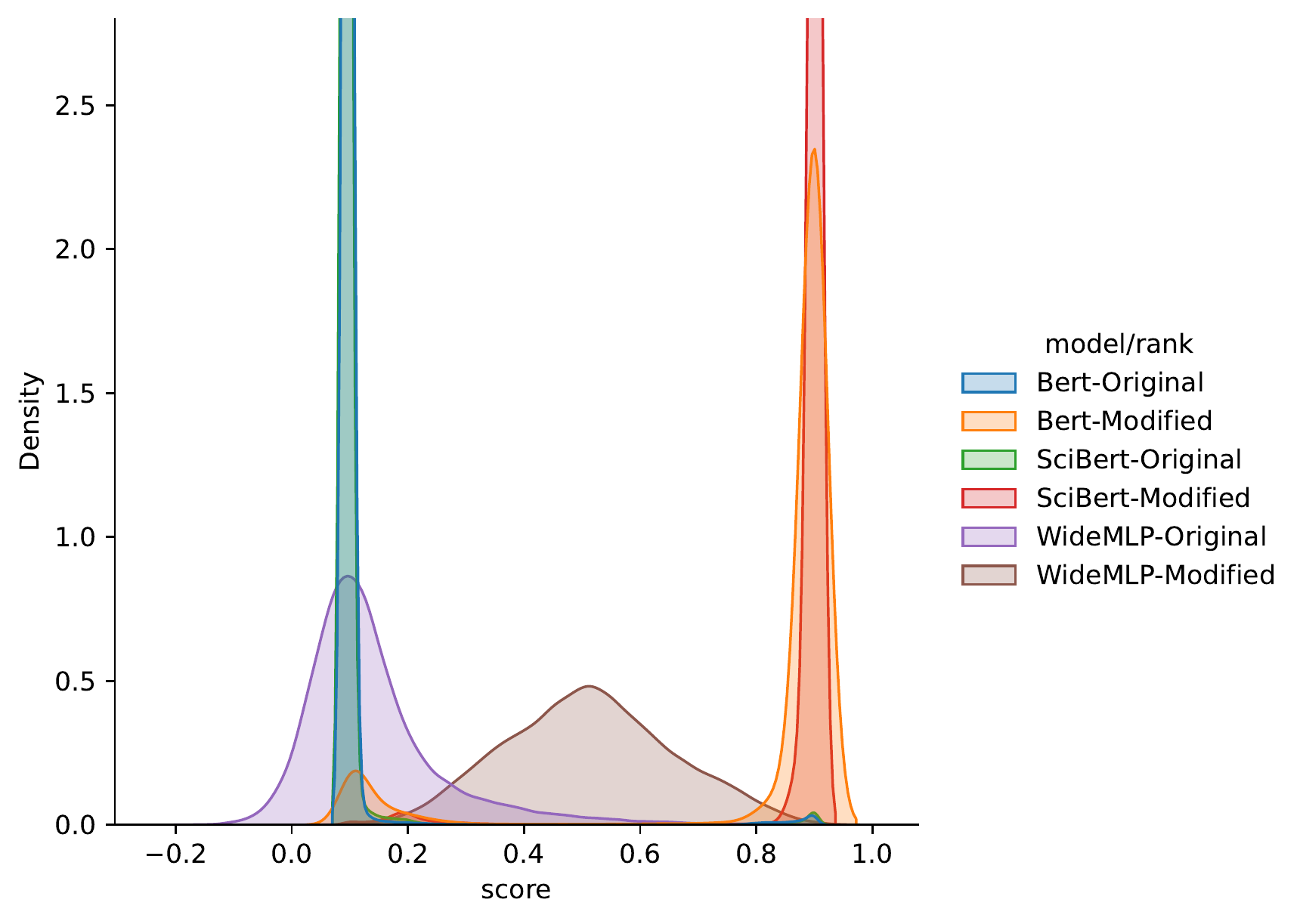}
\caption{Scorings for non-scientific sentences with no modifications (original) and the same sentences with \texttt{<equation>} and \texttt{<reference>} tokens being randomly inserted (modified).}
\label{fig:score_token}
\end{figure}

We performed an additional experiment to analyze the influence of the \texttt{<equ\-at\-ion>} and \texttt{<reference>} tokens on non-scientific sentences.
We modified the non-scientific sentences by inserting the specific tokens.
As shown in Figure~\ref{fig:score_token}, Sci\-BERT now scores most non-scientific sentences containing such a token with $0.9$, while BERT still keeps a small amount of sentences with scores in the non-scientific range. 
For WideMLP, we can see that the influence of the tokens is much smaller.
The mean score here is $0.52$, which is lower then in BERT~($0.81$) and Sci\-BERT~($0.88$).
Therefore, WideMLP relies less on these tokens, which makes it more suitable for non-scientific text containing equations.

\subsubsection{Section classification}
For section classification, the WideMLP baseline is consistently outperformed by transformer-based models. 
This might be a result of the lack of of sequence information which the Bag-of-Words approach neglects.
Also, we use pre-trained transformer models but train WideMLP from scratch, which means that the transformer models start with some understanding of language already. 

We observe that the classification performance increases with more context provided (from 1-sentence to 3-sentences).
For example, the BERT model classifies the one-sentence input 
\begin{quote}
"\textit{Then a weighted sum of attention is carried out to get an attended attention over the document for the final predictions.}"
\end{quote}
as possibly fitting into an introduction, related work, or methods. 
However, enhanced by the surrounding sentences to the following input
\begin{quote}
"We present a novel neural architecture, called attention-over-attention reader, to tackle the cloze-style reading comprehension task. \textit{Then a weighted sum of attention is carried out to get an attended attention over the document for the final predictions.} Among several public datasets, our model could give consistent and significant improvements over various state-of-the-art systems by a large margin."
\end{quote}
the sentence is correctly placed in the conclusion.

As shown in Table~\ref{tab:section-results}, the performance is better for sentences from higher-ranked conferences compared to lower-ranked conference.
The fine-tuned Sci\-BERT, which is pre-trained on a scientific corpus, performed slightly better than BERT with no context. 
However, with higher context size, BERT consistently yields the best results. 
Unlike the scientific data from arXiv that we used for fine-tuning, the pre-training corpus of Sci\-BERT mostly comes from the medical domain~\cite{SciBERT}.
Thus, while Sci\-BERT's pre-training on scientific phrasing is beneficial for the small amount of information contained in a single sentence, BERT's more general corpus helps for inputs of two and three sentences.

As shown in Table~\ref{tab:section-results-context}, providing more context to a model during inference improves the performance, even if the model is trained on inputs with less context.
However, the performance improves more if the additional context was already provided during training. 
An exception are transformer models trained on 2-sentences input but tested on 3-sentences inputs: they achieve higher F1-scores than their counterparts trained on 3-sentences inputs.
This shows that providing context helps training, but context during inference is more important.

While the general section classification performance was high, their predictions include label combinations one would not expect to find in a scientific paper.
For example, sentences were assigned no label, more than two labels, or sections that would not typically contain similar sentences (\eg ``introduction'' and ``experiment'').
In pre-experiments, applying individual thresholds per class or limiting the number of assigned labels to one or two labels per sentence affected $<1.21\%$ of outputs and improved the sample-based F1-score only by $<0.01\%$ for the best model.
Therefore, the influence can be neglected.

\subsubsection{Sentence paraphrasing}
Our task differs from general paraphrasing, since we focus specifically on scientific sentence improvement, where we expect that the input is already quite good and only few changes are necessary.
However, most baseline models have higher BERTScores, which means that these paraphrasers can still keep the semantics of the input, which makes them better general-purpose paraphrasers.
We observe that fine-tuning on IDM-DS gives a $7\%$ larger BLEU score than fine-tuning on Pegasus-DS.
GPT-2 with our custom prompt has a low self-BLEU but high BLEU score, which means that it changes the input a lot and that the output is different from the gold standard.
The low performance of GPT-2 may be attributed to the lack of fine-tuning the model. 
Finally, we observe that the amount of changes in the sentences increases with a higher corruption level.
This means that more sentences are changed when the dissimilarity to the original scientific sentence increases.

\subsection{Threats to Validity}
\label{sec:threattovalidity}
We provide a method for distinguishing whether a sentence is scientific or not.
The selection and labeling of the non-scientific datasets may pose a limitation, which could be improved by using a wider range of non-scientific datasets and more fine-grained scientificness scores.
We carefully investigated the influence of \texttt{<equation>} and \texttt{<reference>} tokens. 
Although the experiments showed that the tokens increase the scientificness of a sentence, this is not an issue, since references and citations are in fact indicators of high scientificness.
For the sentence paraphrasing, the output sentence can be equal to the input sentence. 
An unchanged sentence can be a problem for general paraphrasing, where the model should provide a variety of different suggestions.
However, this is not an issue for us, since the input sentence can be already (quite) scientific.

\subsection{Ethical Considerations}
The development of AI systems in fields like scientific writing needs consideration of ethical and social impact.
Common problems of recent language models are authorship and hallucinations~\cite{ChatGPTEthics}.
Our models do not present these concerns. 
The only models we trained that generate text are the paraphrasers, which aim to maintain the meaning of the input sentence without introducing any new information, whether real or fake.
If one were to deploy our models for interactive writing support, users should check the suggested paraphrases and not blindly integrate them into their text.
This applies to all writing support tools, even simple non-AI variants like Overleaf's dictionary that at times may suggest wrong replacements for technical terms or unknown words.
As with other language models, there is a possibility of extracting training samples~\cite{DBLP:conf/uss/CarliniTWJHLRBS21,DBLP:journals/corr/abs-2212-03749}. 
However, the pre-training checkpoints of our generative models are publicly available and we fine-tuned them on public papers only, which should not contain sensitive private information.
In contrast to proprietary language models that may entail high costs, both in training as well as operation, and thus makes some inaccessible, our models are open-source and accessible to anyone.

\section{Conclusion}
\label{sec:conclusion}
While scientific writing remains a complex task, machine learning methods can be leveraged to be of assistance. 
We show that transformer models achieve the best results in computing a score of scientificness for a sentence, classifying a sentence to a section within the structure of a scientific paper, and paraphrasing scientific sentences.
SciBERT, which is pre-trained on a scientific corpus comprising mostly papers from the broad biomedical domain~\cite{SciBERT}, does not outperform the general-purpose BERT model~\cite{BERT} on tasks for scientific texts from the computer science domain.
We also showed that transformer models profit from context during training and evaluation, 
with providing more context during evaluation being more important than providing it during training.

There are also other datasets such as unarXiv~\cite{DBLP:journals/scientometrics/SaierF20} that we considered using.
Due to the lack of providing section information with the text parapgrahs, we created our own section extraction and mapping approach.
In March 2023, an updated unarXive 2022 dataset~\cite{DBLP:journals/corr/abs-2303-14957} was released that provides structured full text, \ie per paragraph the section title, section type, content type, etc.
It would be interesting to repeat the experiments with this dataset that was not available yet at the time of writing. 

\textit{Acknowledgement.} 
This work is co-funded under the 2LIKE project by the German Federal Ministry of Education and Research (BMBF) and the Ministry of Science, Research and the Arts Baden-Württemberg within the funding line Artificial Intelligence in Higher Education.
We thank C. Schindler, D. Podjavorsek, and S. Birkholz for an early version of the section headings synonym dictionary.

\bibliographystyle{splncs04}
\bibliography{mybibliography}

\end{document}